\RequirePackage{fix-cm}
\documentclass[twocolumn]{svjour3}
\pdfoutput=1
\smartqed 
\usepackage{graphicx}
\usepackage{array}
\usepackage{float}
\usepackage[export]{adjustbox}
\usepackage{indentfirst}
\usepackage{amsmath}
\usepackage{hyperref}

\usepackage{latexsym}
\usepackage{times}
\usepackage{epsfig}
\usepackage{epstopdf}
\usepackage{amssymb}
\usepackage{algorithm2e}
\usepackage{gensymb}
\usepackage{xcolor}
\usepackage{cite}
\usepackage{marvosym}
\usepackage{color}
\usepackage{amsmath,amssymb,amsfonts}
\usepackage{algorithmic}
\usepackage{graphicx}
\usepackage{textcomp}
\usepackage{xcolor}
\usepackage[utf8]{inputenc} % allow utf-8 input
\usepackage[T1]{fontenc}    % use 8-bit T1 fonts
\usepackage{hyperref}       % hyperlinks
\usepackage{url}            % simple URL typesetting
\usepackage{microtype} %for correct hyphenation.
\usepackage{tikz} % Diagramlar çəkmək üçün
\usetikzlibrary{shapes, arrows} % Şəkil və oxlar üçün kitabxana
\usepackage{graphicx}
\usetikzlibrary{arrows, positioning}
\begin{document}

\title{LightFFDNets: Lightweight Convolutional Neural Networks for Rapid Facial Forgery Detection %\thanks{Grants or other notes
}


\author{\href{https://orcid.org/}{\includegraphics[scale=0.06]{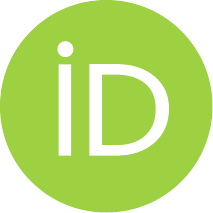}\hspace{1mm}Gu}nel Jabbarli\textsuperscript{1}          \and
        \href{https://orcid.org/0000-0002-3236-5595}{\includegraphics[scale=0.06]{orcid.pdf}\hspace{1mm}Murat Kurt}\textsuperscript{1} %etc.
}


\institute{Gunel Jabbarli \\
           \email{cabbarligunel0899@gmail.com}\\              
             % Tel.: +123-45-678910\\
             % Fax: +123-45-678910\\
                   %  \\
          \at
           Murat Kurt \\            
           \email{murat.kurt@ege.edu.tr}\\
          \at
           \textsuperscript{1} International Computer Institute, Ege University, \.{I}zmir, T{\"u}rkiye
}

\date{Received: date / Accepted: date}

\maketitle

\begin{abstract}
Accurate and fast recognition of forgeries is an issue of great importance in the fields of artificial intelligence, image processing and object detection. Recognition of forgeries of facial imagery is the process of classifying and defining the faces in it by analyzing real-world facial images. This process is usually accomplished by extracting features from an image, using classifier algorithms, and correctly interpreting the results. Recognizing forgeries of facial imagery correctly can encounter many different challenges. For example, factors such as changing lighting conditions, viewing faces from different angles can affect recognition performance, and background complexity and perspective changes in facial images can make accurate recognition difficult. Despite these difficulties, significant progress has been made in the field of forgery detection. Deep learning algorithms, especially Convolutional Neural Networks (CNNs), have significantly improved forgery detection performance.

This study focuses on image processing-based forgery detection using Fake-Vs-Real-Faces (Hard) \cite{hardfakevsrealfaces} and 140k Real and Fake Faces \cite{realandfakefaces} data sets. Both data sets consist of two classes containing real and fake facial images. In our study, two lightweight deep learning models are proposed to conduct forgery detection using these images. Additionally, 8 different pretrained CNN architectures were tested on both data sets and the results were compared with newly developed lightweight CNN models. It’s shown that the proposed lightweight deep learning models have minimum number of layers. It’s also shown that the proposed lightweight deep learning models detect forgeries of facial imagery accurately, and computationally efficiently. Although the data set consists only of face images, the developed models can also be used in other two-class object recognition problems.

\end{abstract}

\section{Introduction}
\label{intro}
Today, the internet is filled with a wealth of visual and video content. This situation is driving the development of search applications and algorithms capable of performing semantic analysis on image and video content to provide users with better search results and summaries. When we see an image, expressing what or who is in it takes less than a millisecond. For instant object recognition, a sequence of information forms within the human brain. Various advancements have been made to replicate this human visual system in computers. Furthermore, object recognition is considered a crucial skill for most computer and robot vision systems. The recognition of objects finds applications in various fields such as autonomous vehicles, automatic forgery detection systems, virtual reality, medical imaging, and e-commerce. Across the globe, different researchers have reported significant progress in areas like image labeling, object detection, and object classification ~\cite{GarciaGarcia2017,Li2010}, enabling the development of approaches to tackle object detection and classification problems.

Deep learning algorithms have revolutionized the field of artificial intelligence and are used especially effectively in image processing ~\cite{Srinivas2016}. CNNs, in particular, have demonstrated remarkable performance in object detection and classification tasks ~\cite{Zhou2014,Wang2014}. CNNs consist of sequential layers that automatically extract features from data and are employed efficiently and effectively in various pattern and image recognition applications, such as gesture recognition ~\cite{Bobic2016}, facial recognition ~\cite{Lawrence1997}, forgery detection~\cite{rosleretal2019} and object classification ~\cite{Guo2016}. Feature extraction is a critical step for these algorithms, involving the derivation of a minimal set of features from images that encapsulate a wealth of object information and capture distinctions between object categories. Deep learning algorithms have successfully addressed complex image processing challenges by utilizing artificial neural networks. These advancements have significantly contributed to some of the most demanding tasks in fields like artificial intelligence, natural language processing, and image processing.

In recent times, machine learning algorithms have become widely prevalent in photo editing applications. These algorithms assist users in image creation, editing, and synthesis while enhancing image quality. As a result, even individuals without expertise in photo editing can generate sophisticated and high-quality images. Furthermore, many photo editing tools and applications offer several intriguing features, such as face swapping, to attract users. Face swapping applications automatically detect faces in photos and allow individuals to replace one person's face with another person's or an animal's face. Face swapping is commonly used on social media and the internet; however, altering faces for malicious purposes or creating fake faces can cause discomfort and make individuals feel uneasy. 

Facial recognition systems typically rely on algorithms based on deep learning and artificial intelligence techniques. These algorithms are trained to recognize new faces based on real face samples stored in a database. However, recent technologies like Generative Adversarial Networks (GANs) have emerged, enabling the creation of realistic fake faces~\cite{Yadavetal2024}. GAN technology, through competition between two neural networks (generator and discriminator), aims to produce high-quality outputs that resemble original inputs. GANs are widely used to generate lifelike new images and enhance existing ones. However, algorithms like GANs can be abused to generate fake information with the intent to deceive people. For example, fake faces created by GANs can be so sophisticated that they not only deceive individuals but can also fool facial recognition systems. Therefore, the problem of distinguishing real from fake facial recognition has become of significant importance today. Facial recognition technology plays a critical role in various areas such as security, privacy, ethics, legal requirements, and artificial intelligence development. The accurate differentiation between real and fake facial images is necessary to address these issues effectively. Fake facial images can be used for purposes like bypassing security systems, engaging in identity theft, or creating deceptive content, jeopardizing individuals' security, privacy, and reputation. Moreover, the correct recognition of real and fake faces is vital for training and performance evaluation of artificial intelligence algorithms. Hence, facial forgery detection is a crucial step for the safety, privacy, and justice of both society and individuals. Different approaches and methods, including deep learning techniques, artificial intelligence algorithms, and image analysis techniques, are being developed to address this challenge. These technologies can perform processes like feature extraction, comparison, and classification to detect differences between real and fake facial images.

In this study, the Fake-Vs-Real-Faces (Hard) \cite{hardfakevsrealfaces} and 140k Real and Fake Faces \cite{realandfakefaces} data sets available in the literature were utilized. In the Fake-Vs-Real-Faces (Hard) data set, containing fake and real images, the images were first resized and then divided into training, testing, and validation sets. Due to the large size of the second data set, a portion of it was selected to match the size of the first data set. Deep learning models based on the Convolutional Neural Network (CNN) architecture were implemented to recognize forgeries in the facial images. Additionally, the newly developed models were compared with existing CNN architectures in the literature, including VGG-16, VGG-19, ResNet-50, ResNet-101, AlexNet, DarkNet-53, MobileNet-V2, and GoogleNet. Although the data set consists solely of facial images, the developed models can also be applied to other two-class object recognition problems.

In summary, the main contributions of this paper are:
\begin{itemize}
    \item  Two novel CNN models (LightFFDNets) for detecting forgeries of facial imagery. 
    %\item A novel integration of residual network.
    \item A detailed validation of our deep learning models (LightFFDNets) on the Fake-Vs-Real-Faces (Hard) \cite{hardfakevsrealfaces} and 140k Real and Fake Faces \cite{realandfakefaces} data sets.
   \item A comparison to the state-of-the-art, showing significant improvements in terms of computation times of facial forgery detection.
%\item A method to learn reparameterizations
%\item A method to learn reparameterizations
\end{itemize}

The second section of this study reviews studies based on object recognition, facial image generation and facial forgery detection, utilizing deep learning techniques based on a CNN architecture. The third section presents our proposed facial forgery detection methods. The fourth section covers the experimental studies conducted as part of this research, presenting the obtained results and their evaluation. Finally, in the fifth and concluding section, a comprehensive assessment of the obtained results is presented, along with a discussion of recommendations for future research.   

\section{Related work}
\label{sec:2}

Our work is related to deep learning models for object detection, facial image generation and detection, so we briefly review each of these below.

\subsection{Object detection}
\label{sec:2.1}

Object detection, one of the most fundamental and challenging problems in the field of computer vision, has attracted great attention in recent years. As noted by Zou et al.~\cite{Zou2023}, comprehensively examine this fast-moving field of research in light of technical evolution spanning more than a quarter of a century (from the 1990s to 2022). In their work, they covered topics such as past landmark detectors, detection datasets, measurements, basic building blocks of the detection system, acceleration techniques and state-of-the-art detection methods.

\subsection{Facial image generation}
\label{sec:2.2}

Rössler et al.~\cite{rosleretal2019} proposed FaceForensics++, which is a dataset of facial forgeries that enables researchers to train deep-learning-based approaches in a supervised fashion. The dataset contains manipulations created with four state-of-the-art methods, namely, Face2Face, FaceSwap, DeepFakes, and NeuralTextures.

Kammoun et al.~\cite{Kammoun2022} reviewed face (Generative Adversarial Networks) GANs and their different applications in their study. They mainly focused on architectures, problems and performance evaluation according to each application and the data sets used. More precisely, they reviewed the progression of architectures and discussed the contributions and limitations of each. Then, they revealed the problems encountered in face GANs and proposed solutions to solve them.

Recently, Yadav et al.~\cite{Yadavetal2024} proposed ISA-GAN, an inception-based self-attentive encoder–decoder network for face synthesis using delineated facial images. Motivated by the reduced computational requirement of the inception network and improved performance and faster convergence by attention networks, Yadav et al.~\cite{Yadavetal2024} utilized the inception, self-attention, and spatial attention modules in the ISA-GAN. Authors showed that the ISA-GAN shows on an average improvement of $9.95\%$ in SSIM score over CUHK dataset while $10.38\%$, $12.58\%$ for WHU-IIP and CVBL-CHILD datasets, respectively.

\subsection{Facial image detection}
\label{sec:2.2}

Soylemez and Ergen~\cite{Soylemez2020} examined the performance of different Convolutional Neural Network architectures in the field of facial expression analysis. In this study, they used the FER2013 dataset. In the study, VGG-16 network was used first. As a result of the training process, the network overfitted the training data, a performance of around $25\%$ was achieved, and the training failed. Inception-V1 showed $65.8\%$ success, Inception-V3 showed $63.2\%$ success, Xception showed $61.1\%$ success, ResNet50-V1 showed $59.5\%$ success, ResNet50-V2 showed $59.8\%$ success, MobileNet-V1, MobileNet-V2 showed $58.5\%$ success.

Adjabi et al.~\cite{Adjabi2020} in their review, they present the history of facial recognition technology, current state-of-the-art methodologies, and future directions. In particular, the newest databases focus on 2D and 3D facial recognition methods. Here, $23$ well-known face recognition datasets are presented in addition to evaluation protocols. Approximately $180$ scientific publications from 1990 to 2020 on facial recognition and its important problems in data collection and preprocessing were reviewed and summarized.

Wang et al.~\cite{Wang2022} studied methods that can detect facial images created or synthesized from GAN models. Existing perception studies are divided into four categories: (1) deep learning-based, (2) physically-based, (3) physiological-based methods, and (4) evaluation and comparison based on human visual performance. For each category, the main ideas are summarized and these are related to method applications.

Alrimy et al.~\cite{Alrimy2022} used three CNN-based methods with VGG-16, Inception-V3, and Resnet50-V2 network architectures to classify facial expressions into seven emotion classes. The facial expression dataset from Kaggle and JAFFE dataset was used to compare the accuracy between the three architectures to find the pre-trained network that best classifies the models. The results showed that the VGG-16 network architecture produced higher accuracy ($93\%$ on JAFFE and $54\%$ on Kaggle) than other architectures.

Lightweight Convolutional Neural Networks were used by Şafak and Barışçı~\cite{Shafak2022} to detect fake face images produced by the whole face synthesis manipulation method. MobileNet, MobileNet-V2, EfficientNet-B0 and NASNetMobile algorithms were used for the training process. The dataset used includes $70,000$ real images from the FFHQ dataset and $70,000$ fake images produced with StyleGAN2 using the FFHQ dataset. In the training process, the weights of the models trained on the ImageNet dataset were reused with transfer learning. The highest accuracy rate was achieved in the EfficientNet-B0 algorithm with a success rate of $93.64\%$.

Nowroozi et al.~\cite{Nowroozi2023} verified synthetic face images using a large printed document dataset. They presented a new dataset consisting of multiple synthetic and natural printed irises taken from VIPPrint Printed and Scanned facial images.

Hamid et al.~\cite{Hamid2023} proposed a computer vision model based on Convolutional Neural Networks for fake image detection. A comparative analysis of 6 popular traditional machine learning models and 6 different CNN architectures was conducted to select the best possible model for further experiments. The proposed model based on ResNet-50 used with powerful preprocessing techniques resulted in an excellent fake image detector with an overall accuracy of $0.99$, with a performance improvement of approximately $18\%$ over other models.

Rodrigo et al.~\cite{Rodrigoetal2024} conducted a thorough comparison between Vision Transformers (ViTs) and CNNs in the context of face recognition tasks, including face identification and face verification subtasks. Their results show that ViTs outperform CNNs in terms of accuracy and robustness against distance and occlusions for face recognition related tasks, while also presenting a smaller memory footprint and an impressive inference speed, rivaling even the fastest CNNs.

%Yun et al.~\cite{Yun2023} review machine learning techniques to recognize objects, features and create process plans in their study. They explain the potential of machine learning in object or feature recognition and provide insight into its application in various smart manufacturing applications. The study describes machine learning methods frequently used in manufacturing, along with a brief introduction of the basic principles. After reviewing traditional object recognition methods, the paper discusses recent studies and insights on feature recognition and manufacturability analysis using machine learning.

%In their study, Pohtongkam and Srinonchat~\cite{Pohtongkam2023} examined the object learning and recognition system through robot touches by developing an artificial sensory system that acts like an electronic skin with tactile sensors. The Tactile Sensor developed in this research consists of 15 Tactile Sensor Arrays and palm contact points. Additionally, recognition analysis was developed on Bag of Word (BoW) and Convolutional Neural Network algorithms. Using Support Vector Machine as classifier with Moment Analysis Descriptor with BoW technique provided the highest accuracy, showing over $80.15\%$ accuracy from capturing an object five times. With the CNN approach, InceptionNet-V3 achieved the highest accuracy of $98.28\%$ from capturing an object only once.

Our models have distinct differences from previous models: two CNN models have been developed, and these models have achieved the best results with very few layers and a maximum of 10 epochs. Additionally, our models have been compared with 8 different transfer learning models. They have a fast processing speed and have been trained on two different datasets. Their applicability for other binary classification problems further expands their range of use.

Since the models are presented in a sequential manner, they are very easy to understand. For each epoch, three trials were conducted for all models, and the results were calculated based on averages. This approach enhances the reliability of the results and allows for a clearer evaluation of model performance.

\section{Our Lightweight Deep Learning Models}

%\subsection{Introduction}
The deep learning models developed in this article addresses object recognition issues using deep learning and transfer learning methods, focusing on static color images and their features, such as shape and color.

\begin{figure}[h]
    \centering
    \begin{tikzpicture}[
        block/.style={
            rectangle, 
            draw=blue,  % Blue border
            fill=blue!20,  % Light blue fill
            text width=2cm, 
            text centered, 
            minimum height=1cm,
            rounded corners=3pt,  % Slightly rounded corners
            thick  % Thicker border
        },
        line/.style={
            draw=blue,  % Blue arrows
            -latex',
            thick  % Thicker lines
        }
    ]
    % Nodes with different color combinations
    \node [block, fill=yellow!20, draw=orange] (dataset) {Suitable dataset};
    \node [block, fill=green!20, draw=green!60!black, below left=1cm and 0.5cm of dataset] (real) {Real face};
    \node [block, fill=red!20, draw=red!60!black, below right=1cm and 0.5cm of dataset] (fake) {Fake face};
    \node [block, fill=violet!20, draw=violet!60!black, below=2cm of dataset] (cnn) {CNN};
    \node [block, fill=cyan!20, draw=cyan!60!black, below=1cm of cnn] (classification) {Classification\\(Real/Fake)};
    
    % Connections with matching colors
    \path [line, orange] (dataset) -- (real);
    \path [line, orange] (dataset) -- (fake);
    \path [line, violet!60!black] (real) -- (cnn);
    \path [line, violet!60!black] (fake) -- (cnn);
    \path [line, cyan!60!black] (cnn) -- (classification);
    \end{tikzpicture}
    \caption{Flow Chart of the Recommended System.}
    \label{fig:schema}
\end{figure}
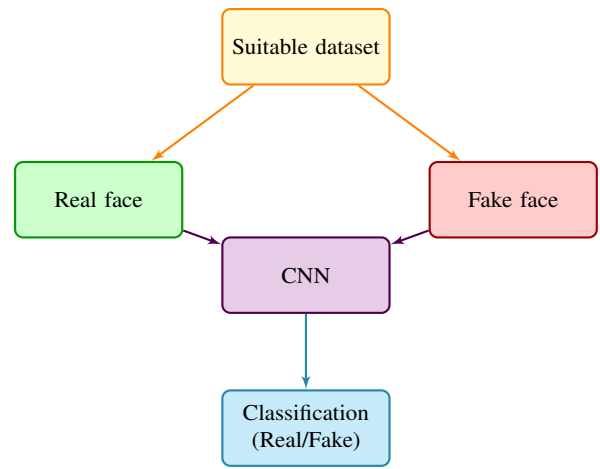

Despite consisting of only a few layers and being trained for a limited number of epochs, the proposed models achieve accuracy levels nearly equivalent to existing models. The reduced number of layers significantly lowers computational complexity, providing a speed advantage over other models. The flowchart of the proposed system is shown in Figure \ref{fig:schema}.

Initially, existing studies on object recognition were reviewed, and suitable datasets were selected. In the next step, preprocessing was performed on selected datasets, which included extracting a subset, dividing the images into training, testing, and validation sets, and resizing all images.

Finally, three different experimental groups were conducted on the identified datasets:

1) Performance values were obtained for a sequential model using facial images.

2) Performance values were obtained using transfer learning with 8 different CNN architectures for selected datasets.

3) The obtained performance values were experimentally compared.

\subsection{Used Datasets}

In this study, two freely available datasets were identified for real and fake face images: Fake-Vs-Real-Faces (Hard) \cite{hardfakevsrealfaces} and 140k Real and Fake Faces \cite{realandfakefaces}. These datasets were obtained from the open-access website for data scientists and machine learning practitioners, Kaggle (https://www.kaggle.com/), without any associated costs. Experimental studies were conducted on these datasets, which contain both real and fake facial images.

\subsubsection{Fake-Vs-Real-Faces (Hard) Dataset}

The Fake-Vs-Real-Faces (Hard) dataset \cite{hardfakevsrealfaces} is curated for the classification task of distinguishing between fake and real human faces. The synthetic faces collected in this dataset were generated using StyleGAN2, making it challenging even for the human eye to accurately classify them. Real human faces, reflecting various characteristics such as age, gender, makeup, ethnicity, etc., were collected to represent diversity encountered in a production environment. This diversity is a crucial factor in evaluating the ability of classification algorithms to recognize real faces accurately under different conditions.

Images in the dataset are in JPEG format and have standard dimensions of $300\times300$ pixels. The "Fake" faces were collected from the website https://thispersondoesnotexist.com. The "Real" face images were obtained through the Unsplash website's API and then cropped using the OpenCV library.

\begin{figure}[h] 
    \centering
    \includegraphics[width=0.4\textwidth]{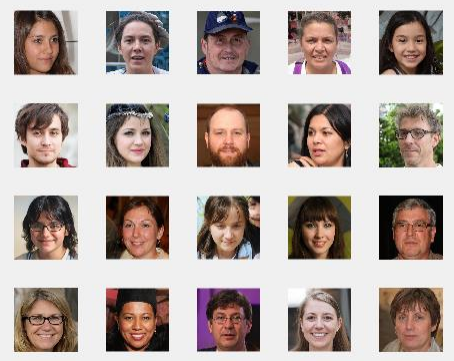} 
    \caption{Sample images from the Fake-Vs-Real-Faces (Hard) dataset \cite{hardfakevsrealfaces}.} 
    \label{fig:sample_images} % Updated label to be descriptive
\end{figure}

The dataset consists of a total of $1288$ images, containing two classes: "Fake" faces ($700$ images) and "Real" faces ($588$ images). Sample images from the dataset are presented in Figure \ref{fig:sample_images}.

\subsubsection{140k Real and Fake Faces Dataset}

This dataset includes $70,000$ REAL faces from the Flickr-Faces-HQ (FFHQ) dataset \cite{Karras2019} collected by Nvidia, as well as $70,000$ FAKE faces sampled from 1 million synthetic faces generated by StyleGAN, provided by Bojan. The FFHQ dataset consists of $70,000$ high-quality PNG images at a resolution of $1024\times1024$, displaying significant variations in age, ethnicity, and image backgrounds. It also covers accessories such as glasses, sunglasses, hats, and others in a comprehensive manner. The images were scraped from Flickr, inheriting all biases of that platform, and were automatically aligned and cropped using dlib. Only images under permissible licenses were collected.

\begin{figure}[h] 
    \centering
    \includegraphics[width=0.4\textwidth]{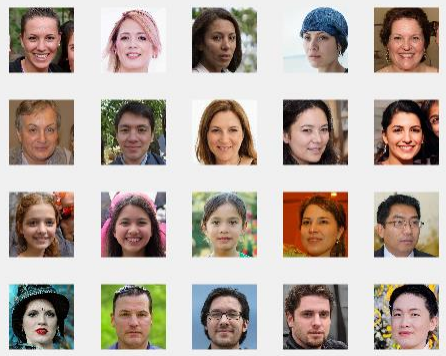} 
    \caption{Sample images from the 140k Real and Fake Faces dataset \cite{realandfakefaces}.} 
    \label{fig:sample_images_140k} % Updated label to be descriptive
\end{figure}

In the 140k Real and Fake Faces dataset \cite{realandfakefaces}, both datasets were properly combined, all images were resized to $256\times256$ pixels, and the data was divided into training, validation, and test sets. Figure \ref{fig:sample_images_140k} presents sample images from the dataset.

\subsection{Our Facial Forgery Detection Models Based on CNN Architecture}
\begin{figure}[t] 
    \centering
    \includegraphics[width=0.4\textwidth]{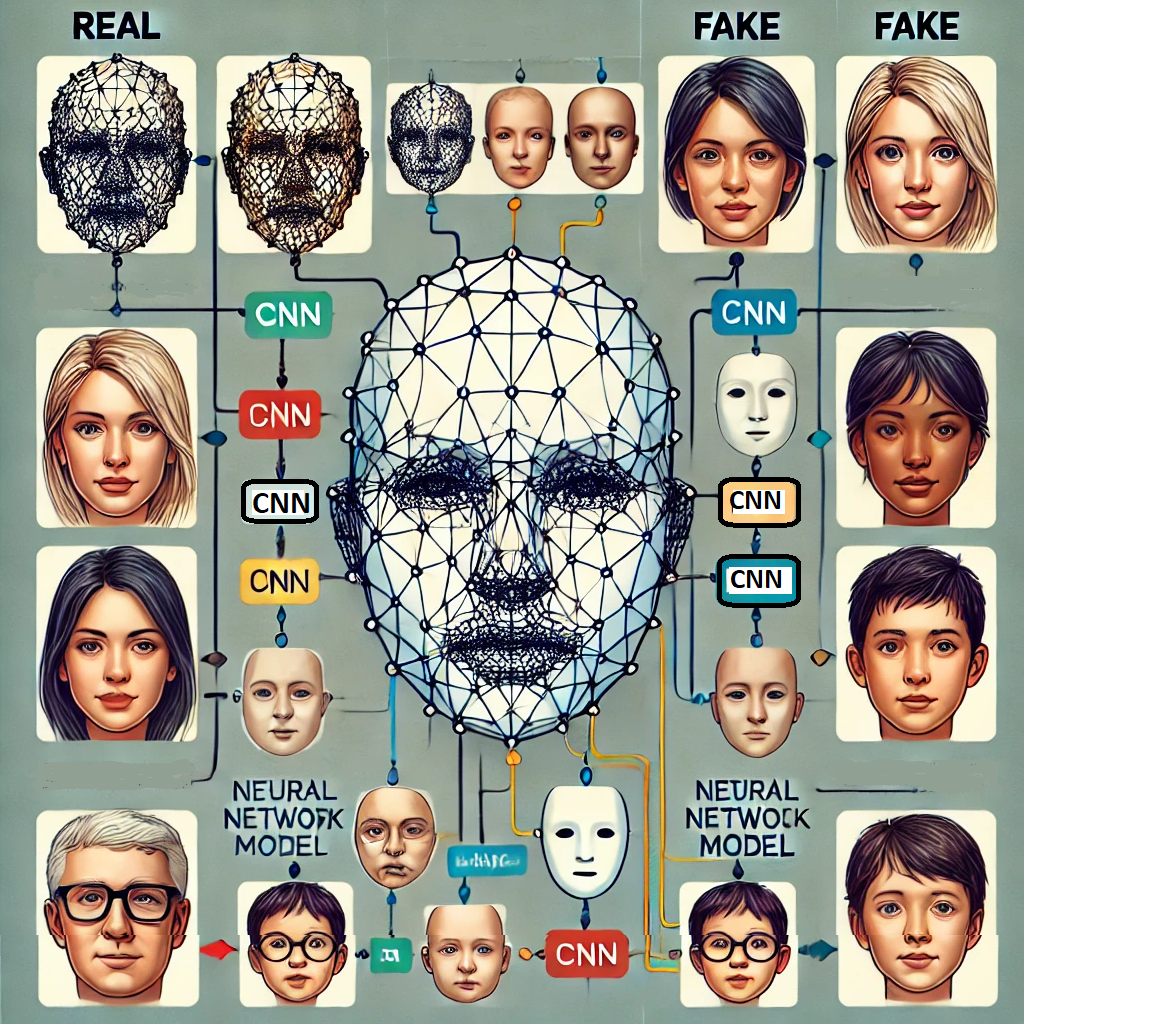} 
    \caption{A general overview of our deep learning models (LightFFDNets) developed in this study.} 
    \label{fig:Model}
\end{figure}

In the literature, different architectures are developed for various problems. While some architectures perform well for a specific problem or dataset, others can achieve successful results in solving many problems. In this study, it is aimed to design an architecture that performs better than existing architectures in two-class (binary) object recognition problems, based on both datasets. Therefore, deep learning models have been developed using the CNN architecture within the scope of this article. An overview of our sequential models (LightFFDNets) is shown in Figure \ref{fig:Model}.

\begin{figure*}[!ht]
    \centering
    \resizebox{\textwidth}{!}{
    \begin{tikzpicture}[
        node distance = 0.5cm,
        box/.style = {
            rectangle, 
            draw=blue!60!black, 
            fill=blue!10,
            rounded corners=2pt,
            minimum width=1.8cm, 
            minimum height=0.6cm,
            text centered,
            font=\footnotesize,
            thick
        },
        arrow/.style = {
            thick,
            ->,
            >=stealth,
            draw=blue!60!black
        }
    ]

    % First column
    \node[box, fill=green!10, draw=green!60!black] (input) {Input Layer};
    \node[box] (conv1) [below=0.3cm of input] {Convolution};
    \node[box, fill=yellow!10, draw=orange] (norm1) [below=0.3cm of conv1] {Normalization};
    \node[box, fill=violet!10, draw=violet!60!black] (act1) [below=0.3cm of norm1] {Activation};
    \node[box, fill=cyan!10, draw=cyan!60!black] (pool1) [below=0.3cm of act1] {Pooling};

    % Second column
    \node[box] (conv2) [right=1.2cm of conv1] {Convolution};
    \node[box, fill=yellow!10, draw=orange] (norm2) [below=0.3cm of conv2] {Normalization};
    \node[box, fill=violet!10, draw=violet!60!black] (act2) [below=0.3cm of norm2] {Activation};
    \node[box, fill=cyan!10, draw=cyan!60!black] (pool2) [below=0.3cm of act2] {Pooling};

    % Third column
    \node[box] (conv3) [right=1.2cm of conv2] {Convolution};
    \node[box, fill=yellow!10, draw=orange] (norm3) [below=0.3cm of conv3] {Normalization};
    \node[box, fill=violet!10, draw=violet!60!black] (act3) [below=0.3cm of norm3] {Activation};
    \node[box, fill=cyan!10, draw=cyan!60!black] (pool3) [below=0.3cm of act3] {Pooling};

    % Fourth column
    \node[box] (conv4) [right=1.2cm of conv3] {Convolution};
    \node[box, fill=yellow!10, draw=orange] (norm4) [below=0.3cm of conv4] {Normalization};
    \node[box, fill=violet!10, draw=violet!60!black] (act4) [below=0.3cm of norm4] {Activation};
    \node[box, fill=cyan!10, draw=cyan!60!black] (pool4) [below=0.3cm of act4] {Pooling};

    % Fifth column
    \node[box] (conv5) [right=1.2cm of conv4] {Convolution};
    \node[box, fill=yellow!10, draw=orange] (norm5) [below=0.3cm of conv5] {Normalization};
    \node[box, fill=violet!10, draw=violet!60!black] (act5) [below=0.3cm of norm5] {Activation};

    % Final layers
    \node[box, fill=red!10, draw=red!60!black] (fc) [right=1.2cm of conv5] {FC Layer};
    \node[box, fill=violet!10, draw=violet!60!black] (actf) [below=0.3cm of fc] {Classification};
    \node[box, fill=green!10, draw=green!60!black] (output) [below=0.3cm of actf] {Output};

    % Vertical connections
    \foreach \i in {1,2,3,4} {
        \path[arrow] (conv\i) -- (norm\i);
        \path[arrow] (norm\i) -- (act\i);
        \path[arrow] (act\i) -- (pool\i);
    }
    \path[arrow] (conv5) -- (norm5);
    \path[arrow] (norm5) -- (act5);

    % Horizontal connections with squared corners
    \path[arrow] (input) -- (conv1);
    \path[arrow] (pool1) -| ([xshift=0.6cm]pool1.east) |- (conv2);
    \path[arrow] (pool2) -| ([xshift=0.6cm]pool2.east) |- (conv3);
    \path[arrow] (pool3) -| ([xshift=0.6cm]pool3.east) |- (conv4);
    \path[arrow] (pool4) -| ([xshift=0.6cm]pool4.east) |- (conv5);
    \path[arrow] (act5) -| ([xshift=0.6cm]act5.east) |- (fc);
    \path[arrow] (fc) -- (actf);
    \path[arrow] (actf) -- (output);

    \end{tikzpicture}
    }
    \caption{Second deep learning model (LightFFDNet v2) developed in the study.} 
    \label{fig:second_model} % Unique label for the second model
\end{figure*}
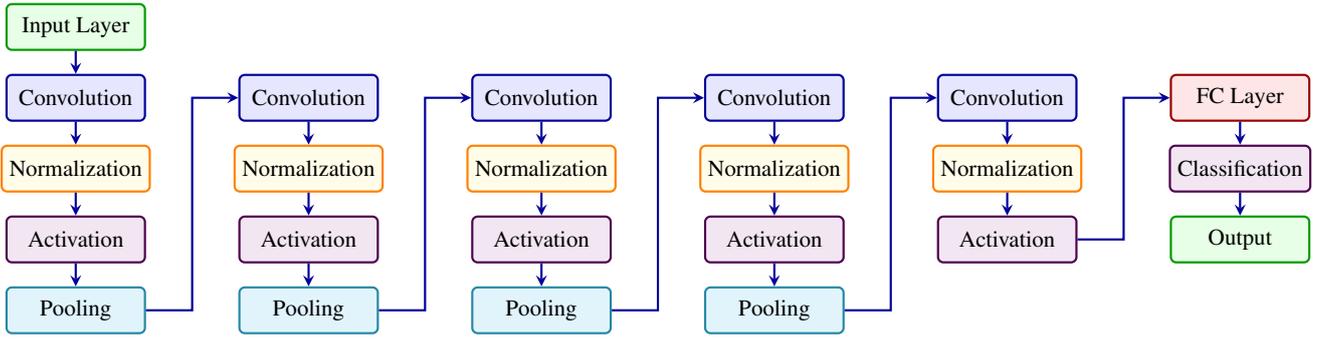

In the first sequential model, named as LightFFDNet v1, 2 convolutional layers and 1 output layer are used (see Figure \ref{fig:first_model}). In the second deep learning model, named as LightFFDNet v2, designed based on the same principle, 5 convolutional layers and 1 output layer are utilized (see Figure \ref{fig:second_model}). In the designed models, images are resized to $224\times224$, and they are provided as input parameters to the input layer. Since classification will be performed in the output layer and there should be as many artificial neural cells as the number of classes, the number of classes has been presented as parameters to the output layer. It should be noted that the number of layers in the models was decided based on the results of the application, by trying out $2, 3, 4, 5$, and $6$ layers to achieve the best results.

\begin{figure}[!ht]
    \centering
    \resizebox{0.5\textwidth}{!}{
    \begin{tikzpicture}[
        node distance = 0.5cm,
        box/.style = {
            rectangle, 
            draw=blue!60!black, 
            fill=blue!10,
            rounded corners=2pt,
            minimum width=1.8cm, 
            minimum height=0.6cm,
            text centered,
            font=\footnotesize,
            thick
        },
        arrow/.style = {
            thick,
            ->,
            >=stealth,
            draw=blue!60!black
        }
    ]

    % First column
    \node[box, fill=green!10, draw=green!60!black] (input) {Input Layer};
    \node[box] (conv1) [below=0.3cm of input] {Convolution};
    \node[box, fill=yellow!10, draw=orange] (norm1) [below=0.3cm of conv1] {Normalization};
    \node[box, fill=violet!10, draw=violet!60!black] (act1) [below=0.3cm of norm1] {Activation};
    \node[box, fill=cyan!10, draw=cyan!60!black] (pool1) [below=0.3cm of act1] {Pooling};

    % Second column
    \node[box] (conv2) [right=1.2cm of conv1] {Convolution};
    \node[box, fill=yellow!10, draw=orange] (norm2) [below=0.3cm of conv2] {Normalization};
    \node[box, fill=violet!10, draw=violet!60!black] (act2) [below=0.3cm of norm2] {Activation};

    % Final layers
    \node[box, fill=red!10, draw=red!60!black] (fc) [right=1.2cm of conv2] {FC Layer};
    \node[box, fill=violet!10, draw=violet!60!black] (actf) [below=0.3cm of fc] {Classification};
    \node[box, fill=green!10, draw=green!60!black] (output) [below=0.3cm of actf] {Output};

    % Vertical connections
    \path[arrow] (conv1) -- (norm1);
    \path[arrow] (norm1) -- (act1);
    \path[arrow] (act1) -- (pool1);
    \path[arrow] (conv2) -- (norm2);
    \path[arrow] (norm2) -- (act2);

    % Horizontal connections with squared corners
    \path[arrow] (input) -- (conv1);
    \path[arrow] (pool1) -| ([xshift=0.6cm]pool1.east) |- (conv2);
    \path[arrow] (act2) -| ([xshift=0.6cm]act2.east) |- (fc);
    \path[arrow] (fc) -- (actf);
    \path[arrow] (actf) -- (output);

    \end{tikzpicture}
    }
    \caption{First deep learning model (LightFFDNet v1) developed in the study.} 
    \label{fig:first_model} % Unique label for the first model
\end{figure}
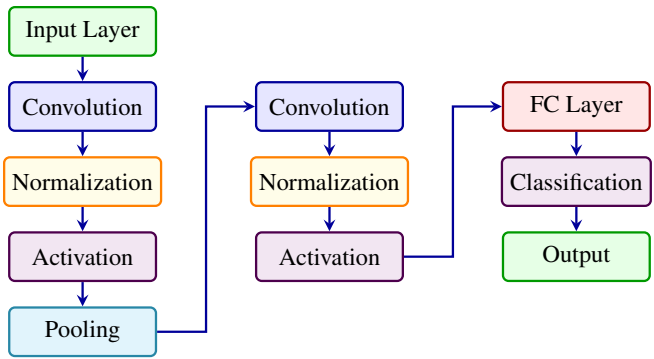

%Model2

In the convolutional layers, $3\times3$ filters were used for feature learning. During this learning process, the filter was applied by sliding over the entire image in each layer. The weights within the filter were multiplied by pixel values and summed to compute the result. As a result of these operations, filters that activated on features detected in the image were learned. The features learned in each layer have a hierarchical structure. Basic features are learned in the initial layers, while higher-level features based on lower-level features are learned in deeper layers.

To reduce the number of parameters in the network, pooling layers are used after the convolutional layers. In these layers, $2\times2$ filters are applied across the image. While the use of filters reduces the number of parameters, max pooling is utilized to retain significant features in the inputs.

The ReLU activation function was selected for the models. This preferred function sets negative values to zero, operates quickly, and has an easily computable derivative. Since classification is performed in this study, the SoftMax activation function was preferred in the output layer to indicate the probability of each input belonging to a class.

The batch normalization used here re-normalizes the weights of all mini-batches in each layer during forward propagation. Thus, the outputs of each mini-batch have zero mean and unit variance. After backpropagation, the updated weights are re-normalized. This way, the outputs from high activation functions are compressed to a suitable range, reducing the risk of overfitting.

In the constructed models (LightFFDNets), the features of the components are combined, passed through convolution, pooling, and activation layers, and then provided as input to the fully connected layer for classification. The models (LightFFDNets) are concluded with the output layer, which is a type of fully connected layer. Since there are 2 classes in the output layer, there are 2 artificial neural cells.

\section{Experimental Studies}\label{futureworks}
%\subsection{Introduction}

In this study, the object recognition system developed is trained and tested on two different datasets. Preprocessing is applied to both datasets (a portion is taken from the dataset, separated into test, training, and validation sets, and all images are resized) for evaluation. 

The study includes three different experimental study groups conducted on the Fake-Vs-Real-Faces (Hard) \cite{hardfakevsrealfaces} and 140k Real and Fake Faces \cite{realandfakefaces} datasets:

\begin{enumerate}
    \item Performance values have been obtained on our deep learning models (LightFFDNets) for face images.
    \item Using the transfer learning method, performance values have been obtained on 8 different CNN architectures for both datasets.
    \item The obtained performance values have been compared.
\end{enumerate}
Our deep learning models have been developed using the MATLAB \cite{matlab} programming language. During the image loading and preprocessing stages, the functions imageDatastore, augmentedImageDatastore, imresize(), inputsize(), numel(), and length() defined in MATLAB have been utilized. The results obtained from the experimental studies have been visualized using the functions disp(), plot(), subplot(), imshow(), confusionmat(), and confusionchart().

\subsection{Data Preprocessing}
In this study, the distribution for the Fake-Vs-Real-Faces (Hard) dataset is set as $70\%$ for training, $10\%$ for validation, and $20\%$ for testing. This partitioning ratio also affects the model's performance. When creating this distribution, the allocation of face images in each category is performed randomly. Additionally, the distribution is made so that an equal number of images are taken from each class, and the same images are used for each distribution (Table \ref{tab:table1}). After the distribution is completed, the images are resized from $300 \times 300$ to $224 \times 224$.
\renewcommand{\arraystretch}{1.5} % Adjusts row height
\begin{table}[ht]
\centering
\caption{The number of Training/Validation/Test face images  for the Fake-Vs-Real-Faces (Hard) dataset\cite{hardfakevsrealfaces}.}
\label{tab:table1} % Added label for referencing
\begin{tabular}{|c|c|}
\hline
\multicolumn{2}{|c|}{\begin{tabular}[c]{@{}c@{}}Total Fake-Vs-Real-Faces (Hard) Dataset Split \\ $70\%$ Training, $10\%$ Validation, $20\%$ Test\end{tabular}} \\ \hline
\multicolumn{2}{|c|}{Training Set (902 Images)} \\ \hline
Real & 412 \\ \hline
Fake & 490 \\ \hline
\multicolumn{2}{|c|}{Validation Set (129 Images)} \\ \hline
Real & 59 \\ \hline
Fake & 70 \\ \hline
\multicolumn{2}{|c|}{Test Set (258 Images)} \\ \hline
Real & 118 \\ \hline
Fake & 140 \\ \hline
\end{tabular}
\end{table}

Since the size of the second dataset (140k Real and Fake Faces) is too large, a portion of it was taken to equalize it with the first dataset before proceeding with the application. As the 140k Real and Fake Faces dataset had already been divided into training, validation, and test sets, no additional split was made here. However, care was taken to ensure that the proportions in each distribution were 70\% for training, 10\% for validation, and 20\% for testing, and that the number of images in each class matched that of the first dataset (Table \ref{tab:table2}).

After this, the images were resized from $256 \times 256$ to $224 \times 224$. The image size was adjusted to fit the layers of various CNN models.

\renewcommand{\arraystretch}{1.5} % Adjusts row height
\begin{table}[ht]
\centering
\caption{The number of Training/Validation/Test Face images  for the 140k Real and Fake faces dataset\cite{realandfakefaces}.}
\label{tab:table2} % Added label for referencing
\begin{tabular}{|c|c|}
\hline
\multicolumn{2}{|c|}{\begin{tabular}[c]{@{}c@{}}Total 140K Real and Fake Faces Dataset Split \\ $70\%$ Training, $10\%$ Validation, $20\%$ Test\end{tabular}} \\ \hline
\multicolumn{2}{|c|}{Training Set (902 Images)} \\ \hline
Real & 451 \\ \hline
Fake & 451 \\ \hline
\multicolumn{2}{|c|}{Validation Set (129 Images)} \\ \hline
Real & 65 \\ \hline
Fake & 65 \\ \hline
\multicolumn{2}{|c|}{Test Set (258 Images)} \\ \hline
Real & 129 \\ \hline
Fake & 129 \\ \hline
\end{tabular}
\end{table}

\subsection{Implementation of Our Deep Learning Models Based on CNN Architecture}
In this study, the deep learning models based on the CNN architecture were developed using the Matlab programming language. The developed models present a sequential structure. In the sequential model structure, there is a single input layer. The other layers are arranged in a consecutive manner, where the output of each layer serves as the input for the next. Within the scope of this study, two different deep learning models were developed using sequential model structures.

The imageInputLayer() function, in which the data dimensions are used as parameters, was utilized for the input layer of the deep learning models based on the developed CNN architecture. For face images, the parameter values were set as $224 \times 224\times 3$. In the convolutional layers, the convolution2dLayer() function, where the filter size and number are used as parameter values, was employed. The filter size was set to $3 \times 3$, and the number of filters was set to $32$. The 'Padding' and 'same' parameters were entered to indicate that padding would be applied to preserve the input data size. In the convolutional layers used in the structure of the models, output of each layer was defined as the input for the subsequent layer.

After each convolutional layer, an activation layer and a normalization layer were added. Except for the last one, a max pooling layer was also added after each convolutional layer. To accelerate the learning process and make the network more stable, the batchNormalizationLayer() function was employed in the normalization layer. The reluLayer() function was added as the activation function due to the preference for the ReLU activation function. In the pooling layer, the maxPooling2dLayer() function, which takes the filter size and stride value as parameters, was used. The default parameter values of a $2\times 2$ filter size and a stride value of 2 were applied. In the output layer, as classification was performed, the number of classes was given as a parameter to the fullyConnectedLayer() function, and the softmaxLayer() function was used for the activation function.

For model design, a cell array named layers, containing a sequence of layers that define the Convolutional Neural Network (CNN) architecture, was created. In the designed models, the number of layers was determined by considering only the convolutional and fully connected layers. The first sequential model (LightFFDNet v1) designed in this study contains a total of 3 layers, consisting of 2 convolutional layers and 1 fully connected layer. The second model (LightFFDNet v2), on the other hand, contains a total of 6 layers, consisting of 5 convolutional layers and 1 fully connected layer.

When designing CNN models, the algorithms or techniques used in the models (LightFFDNets) introduce certain parameters that the designer must decide upon, and these parameters are called hyperparameters \cite{Bischl2023}. The hyperparameters used when constructing the CNN models in this study are the mini-batch size, number of training epochs, cost function, learning rate, and optimization functions.

In this study, an initial configuration was set with a mini-batch size of $32$, a learning rate of $0.001$, the sgdm as the optimization function, and a dropout value of $0.2$. However, after experimentation, it was observed that these settings were not successful and reduced accuracy. After several trials, the optimal hyperparameter values for training the developed models were identified. These values are presented in Table \ref{tab:cnn_hyperparameters}. For the number of epochs, three different values (3, 5, and 10) were used, and the results obtained were utilized in the comparison of the models.

\begin{table}[ht]
\centering
\caption{Hyperparameter values used in CNN models.}
\label{tab:cnn_hyperparameters}
\begin{tabular}{|l|l|}
\hline
\textbf{Hyperparameter} & \textbf{Value} \\ \hline
Loss Function & Crossentropy \\ \hline
Optimization Algorithm & Adam \\ \hline
Mini Batch Size & 16 \\ \hline
Number of Epochs & 3, 5, 10 \\ \hline
Learning Rate & 0.0001 \\ \hline
\end{tabular}
\end{table}

All operations were performed using the MATLAB programming language. The hardware and software infrastructure used, along with their versions, are specified in Table \ref{tab:hardware_software_infrastructure}.

\begin{table}[ht]
\centering
\caption{Hardware and software infrastructure used in this study.}
\label{tab:hardware_software_infrastructure}
\resizebox{0.45\textwidth}{!}{ % Adjust the width of the table
\begin{tabular}{|l|l|}
\hline
\textbf{Hardware and Software} & \textbf{Specifications} \\ \hline
Microprocessor & \parbox[t]{4cm}{AMD Ryzen 7 5800H \\ Radeon Graphics 3.20 GHz} \\ \hline
RAM & \parbox[t]{4cm}{16.0 GB DDR4} \\ \hline
GPU & \parbox[t]{4cm}{NVIDIA RTX 3060 \\ Laptop GPU} \\ \hline
Dedicated Video RAM & \parbox[t]{4cm}{6.0 GB} \\ \hline
Deep Learning Framework & \parbox[t]{4cm}{MATLAB R2023a 64 bit} \\ \hline
Programming Language & \parbox[t]{4cm}{MATLAB} \\ \hline
Operating System & \parbox[t]{4cm}{Windows 11} \\ \hline
System & \parbox[t]{4cm}{64-bit operating system, \\ x64-based processor} \\ \hline
\end{tabular}
}
\end{table}
The procedures carried out in this study were performed on a laptop equipped with the Windows 11 operating system, a Ryzen 7 processor, an Nvidia RTX 3060 graphics card with 6 GB of dedicated memory, and 16 GB of RAM.

\subsection{Application of Pre-trained Deep Neural Networks}
Pre-trained neural networks are deep learning models that have been previously trained on large and complex datasets. These models typically learn high-level features from a general dataset and can subsequently be adapted for different tasks. In this study, CNN architectures were initialized using ImageNet weights and retrained with the images in the dataset. A total of eight different CNN architectures were utilized for classification tasks. These include AlexNet \cite{krizhevsky2012imagenet}, VGG-16 \cite{leonardo2018deep}, VGG-19 \cite{simonyan2014very}, DarkNet-53 \cite{redmon2018yolov3}, GoogleNet \cite{szegedy2015going}, MobileNet-V2 \cite{howard2017mobilenets}, ResNet-50 \cite{he2016deep}, and ResNet-101 \cite{he2016deep}. Each of these architectures was trained separately on the images from both datasets for the classification task, and the results were compared with the newly developed models.

In the training of pre-trained deep learning models, the hyperparameter values used for training the newly developed models were also utilized. The batch size was set to 16, with the ReLU activation function and the Adam optimization algorithm employed. The learning rate was determined to be 0.0001, and the validation frequency was set to 3. Classification was performed for 3, 5, and 10 epochs for each model. Similar to our own models, the Fake-Vs-Real-Faces (Hard) \cite{hardfakevsrealfaces} and 140k Real and Fake Faces \cite{realandfakefaces} datasets were used for the applications here as well. The datasets were divided into $70\%$ training, $10\%$ validation, and $20\%$ test for all classification tasks. The training, validation, and test images used in all models were consisted of the same images. Preprocessing was applied to the images to enhance classification accuracy and to adjust the dimensions to fit the layers of various CNN models. After preprocessing, images of different sizes were scaled to a dimension of $224 \times 224$.

\subsection{Experimental Results and Comparisons on Datasets}
The results obtained from training and testing the proposed object recognition methods on the Fake-Vs-Real-Faces (Hard) \cite{hardfakevsrealfaces} and 140k Real and Fake Faces \cite{realandfakefaces} datasets are presented in this section. This section includes the classification results achieved using face images in the proposed method, employing both newly developed models and pre-trained deep networks, along with a comparison of their performance.

To measure the performance of the models more accurately and sensitively in the experimental studies, the relevant experiments were repeated three times for all models, and the average of these trials was taken. This approach yielded statistically more robust values. Trials for each model were conducted for 3, 5, and 10 epochs.

To evaluate the performance of the proposed models, they were compared with eight pre-trained deep learning models in terms of accuracy and time criteria.

Our first proposed model (LightFFDNet v1), consisting of two convolutional layers, achieved the best result on the Fake-Vs-Real-Faces (Hard) dataset with an average validation accuracy of $99.48\%$ in 5 epochs. In comparison with other models, it demonstrated better performance in terms of accuracy than the VGG-16 and VGG-19 models and was the fastest model, leaving all other models significantly behind in terms of time. As for the test results, it achieved the best outcome with an average test accuracy of $99.74\%$ in 10 epochs. In terms of test accuracy, it fell short of only the VGG-19 architecture by a difference of $0.26\%$.

Our second proposed model (LightFFDNet v2), consisting of five convolutional layers, demonstrated an average validation accuracy of $99.74\%$ on the Fake-Vs-Real-Faces (Hard) dataset in 3 and 5 epochs, falling short by only $0.26\%$ compared to the DarkNet-53, ResNet-50, and ResNet-101 models. However, in terms of speed, it was 8 times faster than DarkNet-53 and ResNet-101, and 5 times faster than ResNet-50. According to the test data, it achieved better results than other models (excluding VGG-19) with an accuracy of $99.87\%$ in both 3 and 10 epochs. In terms of speed, it was slightly behind only the AlexNet architecture.

Our LightFFDNet v1 model exhibited better performance on the 140k Real and Fake Faces dataset with fewer epochs. Specifically, it achieved the best result with an average validation accuracy of $75.64\%$ in 3 epochs. According to these results, it outperformed the AlexNet, VGG-16, and VGG-19 models, and was also the fastest model on this dataset, leaving all other models significantly behind in terms of time. Looking at the test results, it reached the highest accuracy of $71.97\%$ with 5 epochs, demonstrating better performance than the AlexNet, VGG-16, and VGG-19 models.

Our LightFFDNet v2 model achieved the best result on the 140k Real and Fake Faces dataset with a validation accuracy of $76.15\%$ in 10 epochs. Considering all epochs, it outperformed the VGG-16, VGG-19, and AlexNet architectures. The best test result was $71.19\%$. In terms of speed, this model also demonstrated significant superiority over the others. It was twice as fast as the MobileNet-V2 architecture, which showed the closest accuracy result, and 9 to 10 times faster than the DarkNet-53 architecture, which achieved the highest accuracy result. For this dataset, this model also outperformed the AlexNet architecture in terms of speed.

When comparing LightFFDNet v1 and LightFFDNet v2 models, the LightFFDNet v2 model achieved better results in terms of validation accuracy on both datasets, while the LightFFDNet v1 model demonstrated superior performance in terms of speed. The results of all models on the validation and test data for both datasets over 10 epochs are presented in the Table \ref{tab:table5}–\ref{tab:table8}. The time values are calculated in seconds.

\begin{table}[htbp]
\centering
\caption{Results on the Fake-Vs-Real-Faces (Hard) Dataset for 10 epochs.}
\label{tab:table5}
\begin{tabular}{|l|c|c|}
\hline
\textbf{Model} & \textbf{Average Validation Accuracy} & \textbf{Time (s)} \\
\hline
LightFFDNet v1 & 99.22 &  \textbf{66} \\
\hline
LightFFDNet v2 & 99.48 & 88 \\
\hline
VGG-16 & 99.48 & 965 \\
\hline
VGG-19 & 99.74 & 1156 \\
\hline
ResNet-50 &  \textbf{100} & 491 \\
\hline
ResNet-101 & 99.48 & 696 \\
\hline
GoogleNet &  \textbf{100} & 256 \\
\hline
AlexNet & 99.74 & 78 \\
\hline
MobileNet-V2 &  \textbf{100} & 226 \\
\hline
DarkNet-53 &  \textbf{100} & 851 \\
\hline
\end{tabular}
\end{table}

\begin{table}[htbp]
\centering
\caption{Results on the Fake-Vs-Real-Faces (Hard) Dataset for 10 epochs using test data.}
\label{tab:table6}
\label{tab:results_test_data_fake_vs_real_faces}
\begin{tabular}{|l|c|}
\hline
\textbf{Model} & \textbf{Average Test Accuracy} \\
\hline
LightFFDNet v1 & 99.74 \\
\hline
LightFFDNet v2 & 99.87 \\
\hline
VGG-16 & 99.61 \\
\hline
VGG-19 &  \textbf{100} \\
\hline
ResNet-50 & 99.74 \\
\hline
ResNet-101 & 99.61 \\
\hline
GoogleNet & 99.61 \\
\hline
AlexNet & 99.74 \\
\hline
MobileNet-V2 & 99.48 \\
\hline
DarkNet-53 & 99.61 \\
\hline
\end{tabular}
\end{table}

\begin{table}[htbp]
\centering
\caption{Results on the 140k Real and Fake Faces Dataset for 10 epochs.}
\label{tab:table7}
\begin{tabular}{|l|c|c|}
\hline
Model & Average Validation Accuracy & Time \\ 
\hline
LightFFDNet v1 & 70.51 & 94 \\ 
\hline
LightFFDNet v2 & 76.15 &  \textbf{84} \\ 
\hline
VGG-16 & 71.80 & 950 \\ 
\hline
VGG-19 & 61.28 & 1088 \\ 
\hline
ResNet-50 & 87.43 & 443 \\ 
\hline
ResNet-101 & 89.49 & 921 \\ 
\hline
GoogleNet & 80.77 & 144 \\ 
\hline
AlexNet & 77.18 & 112 \\ 
\hline
MobileNet-V2 & 84.62 & 227 \\ 
\hline
DarkNet-53 &  \textbf{93.08} & 851 \\ 
\hline
\end{tabular}
\end{table}

\begin{table}[htbp]
\centering
\caption{Results on the 140k Real and Fake Faces Dataset for 10 epochs using test data.}
\label{tab:table8}
\begin{tabular}{|l|c|}
\hline
Model & Average Test Accuracy \\ 
\hline
LightFFDNet v1 & 69.90 \\ 
\hline
LightFFDNet v2 & 71.19 \\ 
\hline
VGG-16 & 67.96 \\ 
\hline
VGG-19 & 59.95 \\ 
\hline
ResNet-50 & 86.05 \\ 
\hline
ResNet-101 & 84.88 \\ 
\hline
GoogleNet & 74.55 \\ 
\hline
AlexNet & 72.74 \\ 
\hline
MobileNet-V2 & 83.59 \\ 
\hline
DarkNet-53 &  \textbf{92.12} \\ 
\hline
\end{tabular}
\end{table}
Our proposed models have also been compared with pre-trained models based on accuracy criteria, as well as F1 Score, precision, and recall metrics. For all applications conducted on the test dataset, confusion matrices were created for 10 epochs, and the values for F1 Score, precision, and recall were calculated. Confusion matrices for all models on the test datasets have been established for both datasets. These confusion matrices were calculated based on the trial that yielded the best results among three attempts over 10 epochs.
\begin{table}[htbp]
\centering
\caption{F1 Score, Recall and Precision results on the Fake-Vs-Real-Faces (Hard) dataset.}
\label{tab:F1 Fake-Vs-Real-Faces (Hard)}
\begin{tabular}{|l|c|c|c|}
\hline
Model & F1 Score & Recall & Precision \\
\hline
LightFFDNet v1 &  \textbf{1} & \textbf{1} & \textbf{1} \\
\hline
LightFFDNet v2 & \textbf{1} & \textbf{1} & \textbf{1} \\
\hline
VGG-16 & 0.9964 & 1 & 0.9929 \\
\hline
VGG-19 & \textbf{1} & \textbf{1} & \textbf{1} \\
\hline
ResNet-50 & \textbf{1} & \textbf{1} & \textbf{1} \\
\hline
ResNet-101 & 0.9964 & 0.9929 & 1 \\
\hline
GoogleNet & 0.9964 & 0.9929 & 1 \\
\hline
AlexNet & \textbf{1} & \textbf{1} & \textbf{1} \\
\hline
MobileNet-V2 & 0.9929 & 0.9859 & 1 \\
\hline
DarkNet-53 & 0.99640 & 1 & 0.9929 \\
\hline
\end{tabular}
\end{table}

As shown in the Table \ref{tab:F1 Fake-Vs-Real-Faces (Hard)} and Table \ref{tab:F1 140k Real and Fake Faces dataset}, when comparing the models based on F1 Score, recall, and precision results on the Fake-Vs-Real-Faces (Hard) dataset, the proposed models have demonstrated the best results for all three metrics. Additionally, the VGG-19, ResNet-50, and AlexNet models have also achieved maximum results in these metrics.
\begin{table}[htbp]
\centering
\caption{F1 Score, Recall, and Precision results on the 140k Real and Fake Faces dataset.}
\label{tab:F1 140k Real and Fake Faces dataset}
\begin{tabular}{|l|c|c|c|}
\hline
Model & F1 Score & Recall & Precision \\ 
\hline
LightFFDNet v1 & 0.7125 & 0.8511 & 0.6202 \\ \hline
LightFFDNet v2 & 0.7454 & 0.7113 & 0.7829 \\ \hline
VGG-16 & 0.7402 & 0.752 & 0.7287 \\ \hline
VGG-19 & 0.6075 & 0.7647 & 0.5039 \\ \hline
ResNet-50 & 0.9004 & 0.9262 & 0.876 \\ \hline
ResNet-101 & 0.8712 & 0.8519 & 0.8915 \\ \hline
GoogleNet & 0.7737 & 0.731 & 0.8217 \\ \hline
AlexNet & 0.7823 & 0.7465 & 0.8217 \\ \hline
MobileNet-V2 & 0.8605 & 0.8605 & 0.8605 \\ \hline
DarkNet-53 & \textbf{0.9385} & \textbf{0.9313} & \textbf{0.99457} \\ \hline
\end{tabular}
\end{table}

When comparing the models based on the F1 Score, Recall, and Precision results on the 140k Real and Fake Faces dataset, DarkNet-53 achieved the best F1 Score. Our LightFFDNet v1 model demonstrated a Recall of 0.8511, outperforming the VGG-16, VGG-19, AlexNet, and GoogleNet architectures, while trailing slightly behind the other models. In terms of Precision, our LightFFDNet v2 model achieved better results. Among all the models, the best Precision result was attained with DarkNet-53.

\section{Conclusions and Future Work}\label{futureworks}
The results obtained from the experimental analysis presented in the previous section contain several significant insights. The models proposed in this study were tested on two distinct datasets created using images from individuals of various ages, ethnicities, and genders, under different lighting conditions and backgrounds. Our LightFFDNet v1 model is significantly faster than all other models, while also achieving an acceptable level of accuracy, and, in fact, performing better than some models. In the experiments conducted on the Fake-Vs-Real-Faces (Hard) dataset, it is nearly four times faster than the GoogleNet and MobileNet-V2 models, seven times faster than ResNet-50, ten times faster than ResNet-101, thirteen times faster than DarkNet-53, fourteen times faster than VGG-16, and seventeen times faster than VGG-19. In the experiments on the 140k Real and Fake Faces dataset, it is two times faster than the MobileNet-V2 model, nearly five times faster than ResNet-50, nine times faster than DarkNet-53, ten times faster than ResNet-101 and VGG-16, and eleven times faster than VGG-19. Our LightFFDNet v2 model is faster than LightFFDNet v1 model on the 140k Real and Fake Faces dataset, while there is only a slight speed difference between them on the other dataset. In the experiments conducted on the Fake-Vs-Real-Faces (Hard) dataset, it is two times faster than the MobileNet-V2 model, three times faster than GoogleNet, five times faster than ResNet-50, nearly eight times faster than ResNet-101, nine times faster than DarkNet-53, eleven times faster than VGG-16, and thirteen times faster than VGG-19. In the experiments on the 140k Real and Fake Faces dataset, it is nearly three times faster than MobileNet-V2, five times faster than ResNet-50, ten times faster than DarkNet-53, eleven times faster than ResNet-101 and VGG-16, and thirteen times faster than VGG-19.

If we also consider computational complexity, both proposed LightFFDNet models are much more efficient and successful than all existing models. The first proposed model consists of 3 layers, while the second model consists of 6 layers, which makes them lightweight CNN models. Among the pretrained deep learning models, the one with the fewest layers is AlexNet, which has 8 layers. Our LightFFDNet v1 model has nearly 3 times fewer layers than the AlexNet model. This model has 5 times fewer layers than VGG-16, 6 times fewer than VGG-19, 7 times fewer than GoogleNet, 17 times fewer than ResNet-50, DarkNet-53, and MobileNet-V2, and nearly 34 times fewer than ResNet-101. Our LightFFDNet v2 model has 2 times fewer layers than VGG-16 and VGG-19, nearly 3 times fewer than GoogleNet, 6 times fewer than ResNet-50, DarkNet-53, and MobileNet-V2, and 12 times fewer than ResNet-101.

On the 140k Real and Fake Faces dataset, all models, especially the sequential models, did not perform well. Since the proposed models are also of a sequential structure, they exhibited lower accuracy on this dataset compared to the other dataset. 

The results obtained from the experiments reached a maximum of 10 epochs; however, studies in the literature typically employ a higher number of epochs. This characteristic, combined with the fewer number of layers, enables our lightweight models to be significantly faster than other models and to reach conclusions in a shorter time. This quality makes our lightweight models superior compared to others for the datasets used.

In the future, we are planning to upgrade and optimize our LightFFDNets by incorporating state-of-the-art methods and techniques~\cite{Gok2023SIU,Azadvatan2024arXiv,Akdogan2024arXiv}. Subsequent studies will focus on testing the models on different datasets and developing them into a generalized model. Additionally, there are plans to enhance the models not only for binary classification datasets but also for problems with multiple classes.

Furthermore, we are also interested in implementing our novel LightFFDNet models for representing Bidirectional Reflectance Distribution Functions (BRDFs)~\cite{Ozturk2006EGUK,Kurt2007MScThesis,Ozturk2008CG,Kurt2008SIGGRAPHCG,Kurt2009SIGGRAPHCG,Kurt2010SIGGRAPHCG,Ozturk2010GraphiCon,Szecsi2010SCCG,Ozturk2010CGF,Bigili2011CGF,Bilgili2012SCCG,Ergun2012SCCG,Toral2014SIU,Tongbuasirilai2017ICCVW,Kurt2019DEU,Tongbuasirilai2020TVC,Akleman2024arXiv}, Bidirectional Scattering Distribution Functions (BSDFs)~\cite{WKB12,Ward2014MAM,Kurt2014WLRS,Kurt2016SIGGRAPH,Kurt2017MAM,Kurt2018DEU}, Bidirectional Surface Scattering Reflectance Distribution Functions (BSSRDFs)~\cite{Kurt2013TPCG,Kurt2013EGSR,Kurt2014PhDThesis,Onel2019PL,Kurt2020MAM,Kurt2021TVC,Yildirim2024arXiv} and multi-layered materials~\cite{WKB12,Kurt2016SIGGRAPH,Mir2022DEU} in computer graphics.

%\begin{acknowledgements}
%\end{acknowledgements}


\bibliographystyle{spmpsci}      % mathematics and physical sciences
\bibliography{arXiv24_LightFFDNets_References}   % name your BibTeX data base

\begin{thebibliography}{10}
\providecommand{\url}[1]{{#1}}
\providecommand{\urlprefix}{URL }
\expandafter\ifx\csname urlstyle\endcsname\relax
  \providecommand{\doi}[1]{DOI~\discretionary{}{}{}#1}\else
  \providecommand{\doi}{DOI~\discretionary{}{}{}\begingroup \urlstyle{rm}\Url}\fi

\bibitem{Adjabi2020}
Adjabi, I., Ouahabi, A., Benzaoui, A., Taleb-Ahmed, A.: Past, present, and future of face recognition: A review.
\newblock Electronics \textbf{9}(8), 1188 (2020).
\newblock \doi{10.3390/electronics9081188}.
\newblock \urlprefix\url{https://doi.org/10.3390/electronics9081188}

\bibitem{Akdogan2024arXiv}
{Akdo{\u{g}}an}, A., {Kurt}, M.: Exttnet: A deep learning algorithm for extracting table texts from invoice images.
\newblock arXiv preprint arXiv:2402.02246 arXiv:2402.02246 (2024).
\newblock \doi{10.48550/arXiv.2402.02246}.
\newblock \urlprefix\url{https://doi.org/10.48550/arXiv.2402.02246}

\bibitem{Akleman2024arXiv}
Akleman, E., Kurt, M., Akleman, D., Bruins, G., Deng, S., Subramanian, M.: Hyper-realist rendering: A theoretical framework.
\newblock arXiv preprint arXiv:2401.12853 arXiv:2401.12853 (2024).
\newblock \doi{10.48550/arXiv.2401.12853}.
\newblock \urlprefix\url{https://doi.org/10.48550/arXiv.2401.12853}

\bibitem{Alrimy2022}
Alrimy, T., Alloqmani, A., Alotaibi, A., Aljohani, N., Kammoun, S.: Facial expression recognition based on well-known convnet architectures  (2022)

\bibitem{Azadvatan2024arXiv}
{Azadvatan}, Y., {Kurt}, M.: Melnet: A real-time deep learning algorithm for object detection.
\newblock arXiv preprint arXiv:2401.17972 arXiv:2401.17972 (2024).
\newblock \doi{10.48550/arXiv.2401.17972}.
\newblock \urlprefix\url{https://doi.org/10.48550/arXiv.2401.17972}

\bibitem{Bigili2011CGF}
Bilgili, A., {\"O}zt{\"u}rk, A., Kurt, M.: A general {BRDF} representation based on tensor decomposition.
\newblock Computer Graphics Forum \textbf{30}(8), 2427--2439 (2011)

\bibitem{Bilgili2012SCCG}
Bilgili, A., \"{O}zt\"{u}rk, A., Kurt, M.: Representing brdf by wavelet transformation of pair-copula constructions.
\newblock In: Proceedings of the 28th Spring Conference on Computer Graphics, SCCG '12, pp. 63--69. ACM, New York, NY, USA (2012).
\newblock \doi{10.1145/2448531.2448539}.
\newblock \urlprefix\url{http://doi.acm.org/10.1145/2448531.2448539}

\bibitem{Bischl2023}
Bischl, B., Binder, M., Lang, M., Pielok, T., Richter, J., Coors, S., Lindauer, M.: Hyperparameter optimization: Foundations, algorithms, best practices, and open challenges.
\newblock Wiley Interdisciplinary Reviews: Data Mining and Knowledge Discovery \textbf{13}(2), 1484 (2023)

\bibitem{Bobic2016}
Bobić, V., Tadić, P., Kvaščev, G.: Hand gesture recognition using neural network based techniques.
\newblock In: Neural Networks and Applications (NEUREL), 13th Symposium, pp. 1--4. IEEE (2016)

\bibitem{hardfakevsrealfaces}
Boulahia, H.: Hard fake vs real faces dataset.
\newblock \url{https://www.kaggle.com/datasets/hamzaboulahia/hardfakevsrealfaces} (2023).
\newblock Accessed: 2024-10-23

\bibitem{Ergun2012SCCG}
Ergun, S., Kurt, M., \"{O}zt\"{u}rk, A.: Real-time kd-tree based importance sampling of environment maps.
\newblock In: Proceedings of the 28th Spring Conference on Computer Graphics, SCCG '12, pp. 77--84. ACM, New York, NY, USA (2012).
\newblock \doi{10.1145/2448531.2448541}.
\newblock \urlprefix\url{http://doi.acm.org/10.1145/2448531.2448541}

\bibitem{GarciaGarcia2017}
Garcia-Garcia, A., Orts-Escolano, S., Oprea, S., Villena-Martinez, V., Garcia-Rodriguez, J.: A review on deep learning techniques applied to semantic segmentation.
\newblock arXiv preprint arXiv:1704.06857  (2017).
\newblock \urlprefix\url{https://arxiv.org/abs/1704.06857}

\bibitem{Gok2023SIU}
G{\"{o}}k, G., K{\"{u}}{\c{c}}{\"{u}}k, S., Kurt, M., Tar{\i}, E.: A u-net based segmentation and classification approach over orthophoto maps of archaeological sites.
\newblock In: Proceedings of the IEEE 31st Signal Processing and Communications Applications Conference, SIU '23, pp. 1--4. IEEE, Istanbul, Turkey (2023)

\bibitem{Guo2016}
Guo, Y., Liu, Y., Oerlemans, A., Lao, S., Wu, S., Lew, M.S.: Deep learning for visual understanding: A review.
\newblock Neurocomputing \textbf{187}, 27--48 (2016)

\bibitem{Hamid2023}
Hamid, Y., Elyassami, S., Gulzar, Y., Balasaraswathi, V.R., Habuza, T., Wani, S.: An improvised cnn model for fake image detection.
\newblock International Journal of Information Technology \textbf{15}(1), 5--15 (2023)

\bibitem{he2016deep}
He, K., Zhang, X., Ren, S., Sun, J.: Deep residual learning for image recognition.
\newblock In: Proceedings of the IEEE Conference on Computer Vision and Pattern Recognition (CVPR), pp. 770--778 (2016)

\bibitem{howard2017mobilenets}
Howard, A.G., Zhu, M., Chen, B., Kalenichenko, D., Wang, W., Weyand, T., Andreetto, M., Adam, H.: Mobilenets: Efficient convolutional neural networks for mobile vision applications.
\newblock arXiv preprint arXiv:1704.04861  (2017)

\bibitem{Kammoun2022}
Kammoun, A., Slama, R., Tabia, H., Ouni, T., Abid, M.: Generative adversarial networks for face generation: A survey.
\newblock ACM Computing Surveys \textbf{55}(5), 1--37 (2022)

\bibitem{Karras2019}
Karras, T., Laine, S., Aila, T.: A style-based generator architecture for generative adversarial networks.
\newblock In: Proceedings of the IEEE/CVF Conference on Computer Vision and Pattern Recognition, pp. 4401--4410 (2019)

\bibitem{krizhevsky2012imagenet}
Krizhevsky, A., Sutskever, I., Hinton, G.E.: Imagenet classification with deep convolutional neural networks.
\newblock In: Advances in Neural Information Processing Systems, vol.~25 (2012)

\bibitem{Kurt2007MScThesis}
Kurt, M.: A new illumination model in computer graphics.
\newblock Master's thesis, International Computer Institute, Ege University, Izmir, Turkey (2007).
\newblock 140 pages

\bibitem{Kurt2014PhDThesis}
Kurt, M.: An efficient model for subsurface scattering in translucent materials.
\newblock Ph.D. thesis, International Computer Institute, Ege University, Izmir, Turkey (2014).
\newblock 122 pages

\bibitem{Kurt2014WLRS}
Kurt, M.: Grand challenges in bsdf measurement and modeling.
\newblock The Workshop on Light Redirection and Scatter: Measurement, Modeling, Simulation (2014).
\newblock (Invited Talk)

\bibitem{Kurt2017MAM}
Kurt, M.: {Experimental Analysis of BSDF Models}.
\newblock In: R.~Klein, H.~Rushmeier (eds.) Proceedings of the 5th Eurographics Workshop on Material Appearance Modeling: Issues and Acquisition, MAM '17, pp. 35--39. The Eurographics Association, Helsinki, Finland (2017).
\newblock \doi{10.2312/mam.20171330}.
\newblock \urlprefix\url{https://diglib.eg.org:443/handle/10.2312/mam20171330}

\bibitem{Kurt2018DEU}
Kurt, M.: A survey of bsdf measurements and representations.
\newblock Journal of Science and Engineering \textbf{20}(58), 87--102 (2018)

\bibitem{Kurt2019DEU}
Kurt, M.: Real-time shading with phong brdf model.
\newblock Journal of Science and Engineering \textbf{21}(63), 859--867 (2019)

\bibitem{Kurt2020MAM}
Kurt, M.: {A Genetic Algorithm Based Heterogeneous Subsurface Scattering Representation}.
\newblock In: R.~Klein, H.~Rushmeier (eds.) Proceedings of the 8th Eurographics Workshop on Material Appearance Modeling: Issues and Acquisition, MAM '20, pp. 13--16. The Eurographics Association, London, UK (2020).
\newblock \doi{10.2312/mam.20201140}.
\newblock \urlprefix\url{https://diglib.eg.org/handle/10.2312/mam20201140}

\bibitem{Kurt2021TVC}
Kurt, M.: Gensss: a genetic algorithm for measured subsurface scattering representation.
\newblock The Visual Computer \textbf{37}(2), 307--323 (2021).
\newblock \doi{10.1007/s00371-020-01800-0}.
\newblock \urlprefix\url{https://doi.org/10.1007/s00371-020-01800-0}

\bibitem{Kurt2008SIGGRAPHCG}
Kurt, M., Cinsdikici, M.G.: Representing brdfs using soms and mans.
\newblock SIGGRAPH Computer Graphics \textbf{42}(3), 1--18 (2008).
\newblock \doi{http://doi.acm.org/10.1145/1408626.1408630}

\bibitem{Kurt2009SIGGRAPHCG}
Kurt, M., Edwards, D.: A survey of brdf models for computer graphics.
\newblock SIGGRAPH Computer Graphics \textbf{43}(2), 1--7 (2009).
\newblock \doi{http://doi.acm.org/10.1145/1629216.1629222}

\bibitem{Kurt2013EGSR}
Kurt, M., \"{O}zt\"{u}rk, A.: A heterogeneous subsurface scattering representation based on compact and efficient matrix factorization.
\newblock In: Proceedings of the 24th Eurographics Symposium on Rendering, Posters, EGSR '13. Eurographics Association, Zaragoza, Spain (2013)

\bibitem{Kurt2013TPCG}
Kurt, M., \"{O}zt\"{u}rk, A., Peers, P.: A compact tucker-based factorization model for heterogeneous subsurface scattering.
\newblock In: S.~Czanner, W.~Tang (eds.) Proceedings of the 11th Theory and Practice of Computer Graphics, TPCG '13, pp. 85--92. Eurographics Association, Bath, United Kingdom (2013).
\newblock \doi{10.2312/LocalChapterEvents.TPCG.TPCG13.085-092}

\bibitem{Kurt2010SIGGRAPHCG}
Kurt, M., Szirmay-Kalos, L., K\v{r}iv\'{a}nek, J.: An anisotropic brdf model for fitting and monte carlo rendering.
\newblock SIGGRAPH Computer Graphics \textbf{44}(1), 1--15 (2010).
\newblock \doi{http://doi.acm.org/10.1145/1722991.1722996}

\bibitem{Kurt2016SIGGRAPH}
Kurt, M., Ward, G., Bonneel, N.: A data-driven bsdf framework.
\newblock In: Proceedings of the ACM SIGGRAPH 2016, Posters, SIGGRAPH '16, pp. 31:1--31:2. ACM, New York, NY, USA (2016).
\newblock \doi{10.1145/2945078.2945109}.
\newblock \urlprefix\url{http://doi.acm.org/10.1145/2945078.2945109}

\bibitem{Lawrence1997}
Lawrence, S., Giles, C.L., Tsoi, A.C., Back, A.D.: Face recognition: A convolutional neural-network approach.
\newblock IEEE Transactions on Neural Networks \textbf{8}(1), 98--113 (1997)

\bibitem{leonardo2018deep}
Leonardo, M.M., Carvalho, T.J., Rezende, E., Zucchi, R., Faria, F.A.: Deep feature-based classifiers for fruit fly identification (diptera: Tephritidae).
\newblock In: 2018 31st SIBGRAPI Conference on Graphics, Patterns and Images (SIBGRAPI), pp. 41--47. IEEE (2018)

\bibitem{Li2010}
Li, L.J., Su, H., Lim, Y., Fei-Fei, L.: Objects as attributes for scene classification.
\newblock In: Workshop on PaA at European Conference on Computer Vision (ECCV). Heraklion, Crete, Greece (2010)

\bibitem{Mir2022DEU}
Mir, S., Y{\i}ld{\i}r{\i}m, B., Kurt, M.: An analysis of goniochromatic and sparkle effects on multi-layered materials.
\newblock Journal of Science and Engineering \textbf{24}(72), 737--746 (2022)

\bibitem{Nowroozi2023}
Nowroozi, E., Habibi, Y., Conti, M.: Spritz-ps: Validation of synthetic face images using a large dataset of printed documents.
\newblock arXiv preprint arXiv:2304.02982  (2023)

\bibitem{Onel2019PL}
\"{O}nel, S., Kurt, M., \"{O}zt\"{u}rk, A.: An efficient plugin for representing heterogeneous translucent materials.
\newblock In: V.~\.{I}\c{s}ler, H.~G\"{u}r\c{c}ay, H.K. S\"{u}her, G.~\c{C}atak (eds.) Contemporary Topics in Computer Graphics and Games: Selected Papers from the Eurasia Graphics Conference Series, chap.~18, pp. 309--321. Peter Lang GmbH, Internationaler Verlag der Wissenschaften (2019).
\newblock (Book Chapter)

\bibitem{Ozturk2006EGUK}
{\"O}zt{\"u}rk, A., Bilgili, A., Kurt, M.: Polynomial approximation of blinn-phong model.
\newblock In: L.M. Lever, M.~McDerby (eds.) Proceedings of the 4th Theory and Practice of Computer Graphics, TPCG '06, pp. 55--61. Eurographics Association, Middlesbrough, United Kingdom (2006).
\newblock \doi{10.2312/LocalChapterEvents/TPCG/TPCG06/055-061}

\bibitem{Ozturk2010CGF}
{\"O}zt{\"u}rk, A., Kurt, M., Bilgili, A.: A copula-based brdf model.
\newblock Computer Graphics Forum \textbf{29}(6), 1795--1806 (2010)

\bibitem{Ozturk2010GraphiCon}
{\"O}zt{\"u}rk, A., Kurt, M., Bilgili, A.: Modeling brdf by a probability distribution.
\newblock In: Proceedings of the 20th International Conference on Computer Graphics and Vision, pp. 57--63. St. Petersburg, Russia (2010)

\bibitem{Ozturk2008CG}
Ozturk, A., Kurt, M., Bilgili, A., Gungor, C.: Linear approximation of bidirectional reflectance distribution functions.
\newblock Computers \& Graphics \textbf{32}(2), 149--158 (2008)

\bibitem{redmon2018yolov3}
Redmon, J., Farhadi, A.: Yolov3: An incremental improvement.
\newblock arXiv preprint arXiv:1804.02767  (2018)

\bibitem{Rodrigoetal2024}
Rodrigo, M., Cuevas, C., García, N.: Comprehensive comparison between vision transformers and convolutional neural networks for face recognition tasks.
\newblock Scientific Reports \textbf{14} (2024).
\newblock \doi{10.1038/s41598-024-72254-w}

\bibitem{rosleretal2019}
Rossler, A., Cozzolino, D., Verdoliva, L., Riess, C., Thies, J., Niessner, M.: Faceforensics++: Learning to detect manipulated facial images.
\newblock In: 2019 IEEE/CVF International Conference on Computer Vision (ICCV), pp. 1--11. IEEE Computer Society, Los Alamitos, CA, USA (2019).
\newblock \doi{10.1109/ICCV.2019.00009}.
\newblock \urlprefix\url{https://doi.ieeecomputersociety.org/10.1109/ICCV.2019.00009}

\bibitem{simonyan2014very}
Simonyan, K., Zisserman, A.: Very deep convolutional networks for large-scale image recognition.
\newblock arXiv preprint arXiv:1409.1556  (2014)

\bibitem{Soylemez2020}
S\"{o}ylemez, O.F., Ergen, B.: Farklı evrişimsel sinir ağı mimarilerinin yüz İfade analizi alanındaki başarımlarının İncelenmesi.
\newblock Dicle Üniversitesi Mühendislik Fakültesi Mühendislik Dergisi \textbf{11}(1), 123--133 (2020)

\bibitem{Srinivas2016}
Srinivas, S., Sarvadevabhatla, R.K., Mopuri, K.R., Prabhu, N., Kruthiventi, S.S., Babu, R.V.: A taxonomy of deep convolutional neural nets for computer vision.
\newblock Frontiers in Robotics and AI \textbf{2}, 36 (2016)

\bibitem{Szecsi2010SCCG}
Sz\'{e}csi, L., Szirmay-Kalos, L., Kurt, M., Cs\'{e}bfalvi, B.: Adaptive sampling for environment mapping.
\newblock In: Proceedings of the 26th Spring Conference on Computer Graphics, SCCG '10, pp. 69--76. ACM, New York, NY, USA (2010).
\newblock \doi{http://doi.acm.org/10.1145/1925059.1925073}.
\newblock \urlprefix\url{http://doi.acm.org/10.1145/1925059.1925073}

\bibitem{szegedy2015going}
Szegedy, C., Liu, W., Jia, Y., Sermanet, P., Reed, S., Anguelov, D., Rabinovich, A.: Going deeper with convolutions.
\newblock In: Proceedings of The IEEE Conference on Computer Vision and Pattern Recognition, pp. 1--9 (2015)

\bibitem{matlab}
The~MathWorks, I.: Matlab r2023a (2023).
\newblock \urlprefix\url{https://www.mathworks.com}.
\newblock Accessed: 2024-11-07

\bibitem{Tongbuasirilai2020TVC}
Tongbuasirilai, T., Unger, J., Kronander, J., Kurt, M.: Compact and intuitive data-driven brdf models.
\newblock The Visual Computer \textbf{36}(4), 855--872 (2020).
\newblock \doi{10.1007/s00371-019-01664-z}.
\newblock \urlprefix\url{https://doi.org/10.1007/s00371-019-01664-z}

\bibitem{Tongbuasirilai2017ICCVW}
Tongbuasirilai, T., Unger, J., Kurt, M.: Efficient {BRDF} sampling using projected deviation vector parameterization.
\newblock In: Proceedings of the {IEEE} International Conference on Computer Vision Workshops, ICCVW '17, pp. 153--158. {IEEE} Computer Society, Venice, Italy (2017).
\newblock \doi{10.1109/ICCVW.2017.26}.
\newblock \urlprefix\url{http://doi.ieeecomputersociety.org/10.1109/ICCVW.2017.26}

\bibitem{Toral2014SIU}
T\"{o}ral, O.A., Ergun, S., Kurt, M., \"{O}zt\"{u}rk, A.: Mobile gpu-based importance sampling.
\newblock In: Proceedings of the IEEE 22nd Signal Processing and Communications Applications Conference, SIU '14, pp. 510--513. IEEE, Trabzon, Turkey (2014)

\bibitem{Wang2022}
Wang, X., Guo, H., Hu, S., Chang, M.C., Lyu, S.: Gan-generated faces detection: A survey and new perspectives.
\newblock arXiv preprint arXiv:2202.07145  (2022).
\newblock \urlprefix\url{https://doi.org/10.48550/arXiv.2202.07145}

\bibitem{Wang2014}
Wang, Y., Wu, Y.: Scene classification with deep convolutional neural networks.
\newblock Tech. rep., University of California (2014)

\bibitem{WKB12}
Ward, G., Kurt, M., Bonneel, N.: A practical framework for sharing and rendering real-world bidirectional scattering distribution functions.
\newblock Tech. Rep. LBNL-5954E, Lawrence Berkeley National Laboratory (2012)

\bibitem{Ward2014MAM}
Ward, G., Kurt, M., Bonneel, N.: Reducing anisotropic bsdf measurement to common practice.
\newblock In: R.~Klein, H.~Rushmeier (eds.) Proceedings of the 2nd Eurographics Workshop on Material Appearance Modeling: Issues and Acquisition, MAM '14, pp. 5--8. Eurographics Association, Lyon, France (2014).
\newblock \doi{10.2312/mam.20141292}.
\newblock \urlprefix\url{http://diglib.eg.org/EG/DL/WS/MAM/MAM2014/005-008.pdf}

\bibitem{realandfakefaces}
Xhlulu: 140k real and fake faces dataset.
\newblock \url{https://www.kaggle.com/datasets/xhlulu/140k-real-and-fake-faces} (2023).
\newblock Accessed: 2024-10-23

\bibitem{Yadavetal2024}
Yadav, N., Singh, S.K., Dubey, S.R.: Isa-gan: inception-based self-attentive encoder–decoder network for face synthesis using delineated facial images.
\newblock The Visual Computer \textbf{40}, 8205--8225 (2024).
\newblock \doi{10.1007/s00371-023-03233-x}

\bibitem{Yildirim2024arXiv}
{Y{\i}ld{\i}r{\i}m}, B., {Kurt}, M.: Genplusss: A genetic algorithm based plugin for measured subsurface scattering representation.
\newblock arXiv preprint arXiv:2401.15245 arXiv:2401.15245 (2024).
\newblock \doi{10.48550/arXiv.2401.15245}.
\newblock \urlprefix\url{https://doi.org/10.48550/arXiv.2401.15245}

\bibitem{Zhou2014}
Zhou, B., Khosla, A., Lapedriza, A., Oliva, A., Torralba, A.: Object detectors emerge in deep scene cnns.
\newblock arXiv preprint arXiv:1412.6856  (2014).
\newblock \urlprefix\url{https://arxiv.org/abs/1412.6856}

\bibitem{Zou2023}
Zou, Z., Chen, K., Shi, Z., Guo, Y., Ye, J.: Object detection in 20 years: A survey.
\newblock In: Proceedings of the IEEE (2023)

\bibitem{Shafak2022}
Şafak, E., Barişçi, N.: Hafif evrişimsel sinir ağları kullanılarak sahte yüz görüntülerinin tespiti.
\newblock El-Cezeri \textbf{9}(4), 1282--1289 (2022)

\end{thebibliography}

%\begin{thebibliography}{}
% etc
%\end{thebibliography}
\end{document}